\documentclass[11pt,a4paper]{article}
\usepackage{times,latexsym}
\usepackage{url}
\usepackage[T1]{fontenc}
\usepackage{booktabs}
\usepackage{multirow}
\usepackage{graphicx}
\usepackage{tikz}
\usetikzlibrary{arrows.meta, positioning}
\usepackage{dblfloatfix}  
\usepackage{colortbl}
\usepackage{tcolorbox}
\usepackage{amsmath}
\usepackage{makecell} 
\usepackage{float}

\usepackage{xcolor}
\definecolor{apibg}{RGB}{248,248,248}
\newcommand{\zw}{\hspace{0pt}}

\newcommand{\apimapbox}[3]{%
  \noindent\colorbox{apibg}{%
    \parbox{\dimexpr\linewidth-2\fboxsep\relax}{%
      \small
      {\raggedright\hyphenpenalty=10000\exhyphenpenalty=10000
       #1\ \(\rightarrow\)\ \allowbreak #2\par}%
      \vspace{2pt}%
      {\footnotesize\itshape relevance\_score: #3}%
    }%
  }%
}

\newcommand{\apimapboxsim}[3]{%
  \noindent\colorbox{apibg}{%
    \parbox{\dimexpr\linewidth-2\fboxsep\relax}{%
      \small
      {\raggedright\hyphenpenalty=10000\exhyphenpenalty=10000
       #1\ \(\rightarrow\)\ \allowbreak #2\par}%
      \vspace{2pt}%
      {\footnotesize\itshape similarity: #3}%
    }%
  }%
}

\usepackage{listings}
\lstdefinestyle{mystyle}{
    backgroundcolor=\color{gray!10},  % Light gray background
    basicstyle=\ttfamily\scriptsize,  % Monospaced font in small size
    frame=single,                     % Single frame around the code
    framesep=6pt,                     % Separation between frame and code
    framerule=0.5pt,                  % Frame thickness
    rulecolor=\color{black!30},       % Frame color
    breaklines=true,                  % Automatic line breaking
    breakatwhitespace=true,           % Break lines at whitespaces
    breakindent=0pt,                  % No extra indent on wrapped lines
    showstringspaces=false,           % Don't visually mark spaces in strings
    columns=flexible,                 % Flexible column spacing
    captionpos=b,                     % Caption positioned at the bottom
    numberstyle=\tiny\color{gray},     % Style for line numbers
    xleftmargin=10pt,                 % Extra left margin for better readability
    xrightmargin=10pt,                % Extra right margin
    tabsize=4,                        % Set tab size to 4 spaces
    extendedchars=true                % Support for non-ASCII characters
}
\usepackage[acceptedWithA]{tacl2021v1}
\usepackage{xspace,mfirstuc,tabulary}

\newif\iftaclinstructions
\taclinstructionsfalse % AUTHORS: do NOT set this to true
\iftaclinstructions
\renewcommand{\confidential}{}
\renewcommand{\anonsubtext}{(No author info supplied here, for consistency with
TACL-submission anonymization requirements)}
\newcommand{\instr}
\fi

\iftaclpubformat % this "if" is set by the choice of options

\else

\fi

\title{In-N-Out: A Parameter-Level API Graph Dataset for Tool Agents}

\author{
  Seungkyu Lee$^1$\thanks{~~This work was done while the first author was an intern at the Graduate School of Data Science at Seoul National University.} 
  ~~ Nalim Kim$^2$ ~~ Yohan Jo$^2$\thanks{~~Corresponding author.} 
  \\
  $^1$Department of Industrial Engineering, Seoul National University
  \\
  $^2$Graduate School of Data Science, Seoul National University
}

\date{}

\begin{document}
\maketitle
\begin{abstract}
Tool agents---LLM-based systems that interact with external APIs---offer a way to execute real-world tasks. However, as tasks become increasingly complex, these agents struggle to identify and call the correct APIs in the proper order. To tackle this problem, we investigate converting API documentation into a structured API graph that captures API dependencies and leveraging it for multi-tool queries that require compositional API calls. To support this, we introduce \textbf{In-N-Out}, the first expert-annotated dataset of API graphs built from two real-world API benchmarks and their documentation. Using In-N-Out significantly improves performance on both tool retrieval and multi-tool query generation, nearly doubling that of LLMs using documentation alone. Moreover, graphs generated by models fine-tuned on In-N-Out close 90\% of this gap, showing that our dataset helps models learn to comprehend API documentation and parameter relationships. Our findings highlight the promise of using explicit API graphs for tool agents and the utility of In-N-Out as a valuable resource. We release our dataset and code at \url{https://github.com/holi-lab/In-N-Out-API-Graph}.
\end{abstract}

\section{Introduction}
\label{sec1:introduction}

Recent advances in Large Language Models (LLMs) have spurred growing interest in building \textit{tool agents}---LLM-based systems that interact with external APIs to execute real-world tasks \cite{schick2023toolformer, shen2023hugginggpt}. Unlike traditional language models that rely solely on pre-trained knowledge, tool agents can retrieve real-time information, access databases, and operate services via API calls \cite{song2023restgpt}. As user queries grow more complex, solving them often involves chaining multiple APIs---commonly known as \textbf{multi-tool queries}---where one API call requires the outputs produced by a previous call. This makes it crucial for tool agents not only to understand individual APIs but also to capture interdependencies among APIs to plan valid call sequences \cite{lumer2025graphrag, zhang2024reversechain}.

\begin{figure*}[t]
    \centering
    \includegraphics[width=2\columnwidth]{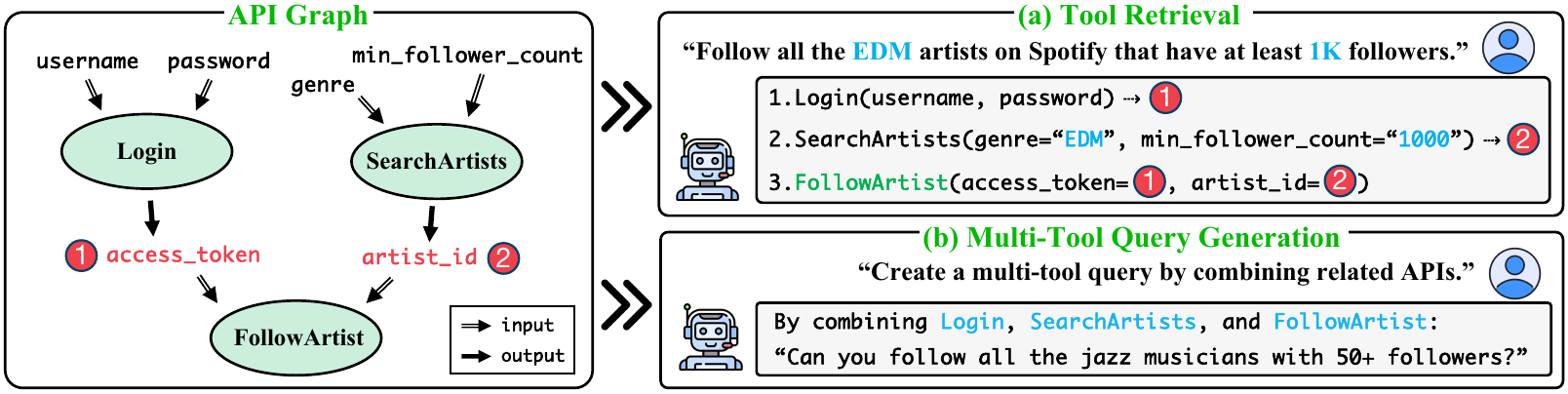}
    \caption{Illustration of how parameter-level API graphs support (a) \textbf{Tool Retrieval}, where the agent identifies prerequisite APIs needed to supply inputs for a target API (e.g., FollowArtist), and (b) \textbf{Multi-Tool Query Generation}, where the agent selects interdependent APIs to construct coherent multi-tool queries.}
    \label{fig:api-graph-usage}
\end{figure*}

A concrete example in Figure~\ref{fig:api-graph-usage}a illustrates why such interdependencies are essential. To fulfill the query \textit{``Follow all the EDM artists on Spotify that have at least 1K followers''}, an agent must first call other APIs to obtain an access token and artist IDs, before invoking FollowArtist(). Without explicit knowledge of parameter connections, agents often fail to chain these APIs correctly, as the required dependencies are obscured among the large number of available APIs.

While prior studies have explored training tool agents on large datasets of multi-tool queries \cite{qin2024toolllm, shim2025tooldial}, a more natural and general solution, akin to what human developers do, is to use API documentation to identify API dependencies. However, since LLMs struggle to comprehend parameter-level dependencies from noisy, real-world API documentation (\S{\ref{sec4:experiments}}), we focus on a relatively underexplored approach: converting API documentation into an explicit API graph, enabling LLMs to use this structured information. Some prior studies that have attempted to use API graphs have focused primarily on limited domains \cite{wang2024solutionbasedllm} or automatic graph construction with heuristics \cite{liu2024controlllm, shen2024taskbench}, leaving the true potential of accurate API graphs in real-world scenarios unexplored. Since manual graph construction is not scalable for arbitrary API sets, a high-quality dataset is needed to train and evaluate a documentation-to-graph conversion module and ensure its generalizability to unseen APIs. Our approach is arguably more generalizable and scalable than simply training a tool agent on large multi-tool tasks, and the graphs can even facilitate the generation of multi-tool queries by providing the underlying API composition for complex queries (Figure~\ref{fig:api-graph-usage}b).

Against this backdrop, we introduce \textbf{In-N-Out}, an API graph dataset constructed through expert annotation. In-N-Out represents both APIs and their parameters as nodes, with directed edges indicating when an output from one API can serve as a valid input to another. The APIs are sourced from AppWorld~\cite{trivedi2024appworld} and NESTful(v1)~\cite{basu2025nestfulbenchmarkevaluatingllms}, spanning 550 APIs, and over 30,000 parameter-level edges, thus covering diverse and realistic tool-use scenarios.

We first demonstrate that training LLMs on our In-N-Out dataset improves their ability to construct parameter-level API graphs from noisy documentation, even for unseen APIs. Specifically, we prompt models to decide whether a connection exists between the parameters of two given APIs. Zero-shot models perform poorly on this task, indicating that LLM-based automatic graph construction is unreliable. In contrast, models fine-tuned on In-N-Out achieve significantly higher accuracy and generalize well to unseen APIs and datasets, sometimes reaching the accuracy achieved by training on the test dataset itself. This demonstrates that In-N-Out allows LLMs to learn the underlying ability to comprehend API documentation and parameter relationships.

We further draw two key insights from experiments on two tasks: (1) solving multi-tool queries and (2) generating multi-tool queries (which may be used for training and benchmarking tool agents). First, using In-N-Out significantly improves performance on both tasks compared to using documentation alone. This justifies our approach of using explicit, high-quality API graphs for overcoming challenges in tool agents. Second, even automatically generated graphs from models fine-tuned on In-N-Out yield substantial gains, closing over 90\% of the performance gap on both tasks compared to using documentation directly. The results demonstrate that In-N-Out can serve as a valuable resource for graph-based approaches for tool agents.

In summary, our contributions are as follows:
\begin{itemize}
    \item We construct and release In-N-Out, an API graph dataset annotated by experts, capturing parameter-level dependencies between real-world APIs.

    \item We show that fine-tuning on In-N-Out enables LLMs to infer parameter-level connections from textual API documentation and generalize to unseen APIs. 

    \item We demonstrate that In-N-Out substantially improve tool agent performance, and that even automatically generated graphs reduce the performance gap relative to human-labeled graphs.
\end{itemize}

\section{Related Work}
\label{sec2:related-work}

Developing tool agents has centered on two key tasks: tool retrieval and planning (\S{\ref{sec2:tool-retrieval-and-planning}}), and multi-tool query generation (\S{\ref{sec2:multi-tool-query-generation}}). Graph-based approaches have supported both tasks by capturing API relationships and enabling scalable coordination.

\subsection{Tool Retrieval and Planning}
\label{sec2:tool-retrieval-and-planning}

As real-world tasks increasingly require multiple APIs in sequence, both accurate retrieval and correct input provision become essential~\citep{patil2025BFCL, xu2024enhancingtoolretrieval, shi2025retrieval, du2024anytool, lumer2024toolshed}. To support these capabilities, previous work has explored several approaches: refining API documentation to make it more interpretable for LLMs~\citep{hsieh2023tooldocumentation, yuan2025easytool, chen2024reinvoke}, fine-tuning LLMs on query-API call pairs~\citep{qin2024toolllm, patil2024gorilla}, and learning tool-specific embeddings~\citep{hao2023toolkengpt, wang2025toolgen}. However, these strategies often fail to capture the underlying dependencies among APIs, while also requiring high training costs and generalizing poorly to unseen APIs.

To address this, recent studies have explored graph-based approaches that represent API relationships explicitly to support tool retrieval~\citep{liu2024controlllm, liu2024toolnet, chen-etal-2025-locagent}. GraphRAG-Tool Fusion~\citep{lumer2025graphrag} constructs graphs by modeling both API- and parameter-level dependencies, but relies on synthetic APIs that lack the complexity of real-world environments. SoAy~\citep{wang2024solutionbasedllm} uses real APIs but is limited to academic information-seeking tasks with only seven APIs and simple parameter structures, falling short of the real-world requirement to combine many APIs across diverse domains. In contrast, we construct expert-verified graphs over real APIs across 25 domains, capturing richer relationships and enabling reliable tool retrieval in real-world settings.

\subsection{Multi-Tool Query Generation}
\label{sec2:multi-tool-query-generation}

Another direction for developing tool agents is to fine-tune them on queries paired with the API call trajectories needed to resolve them, providing direct supervision for both planning and execution~\citep{zhuang2023toolqa, li2023apibank, huang2024metatool}. For such training to be effective, the queries must reflect genuine interdependencies among APIs~\citep{shen2025shortcutsbench, wu2024sealtools}. However, existing efforts face two fundamental challenges: ensuring validity and achieving scalability. NESTful(v1)~\citep{basu2025nestfulbenchmarkevaluatingllms} builds nested queries from inferred relationships with human validation, but its queries are limited in scale. NesTools~\citep{han2025nestools} address scalability by automatically generating queries with LLMs, but relies on synthetic APIs.

To overcome these issues, recent work has leveraged graph structures to systematically generate multi-tool queries~\cite{wang2024solutionbasedllm}. TaskBench~\cite{shen2024taskbench} constructs graphs by matching parameter data types (e.g., image, text), and samples subgraphs to generate queries. ToolDial~\cite{shim2025tooldial} connects APIs through keyword matching and embedding similarity, and generates queries by traversing pairs of connected nodes. While scalable, these approaches often connect semantically incompatible APIs or capture only coarse relations. In contrast, our work constructs parameter-level API graphs through expert annotation, providing accurate ground-truth connections that support valid query generation reflecting genuine dependencies. Moreover, models trained on these graphs can generalize to unseen APIs, enabling scalable inference of new dependencies and corresponding query generation.
\section{In-N-Out Dataset}
\label{sec3:dataset}

\begin{figure*}[t]
  \centering
  \includegraphics[width=\textwidth]{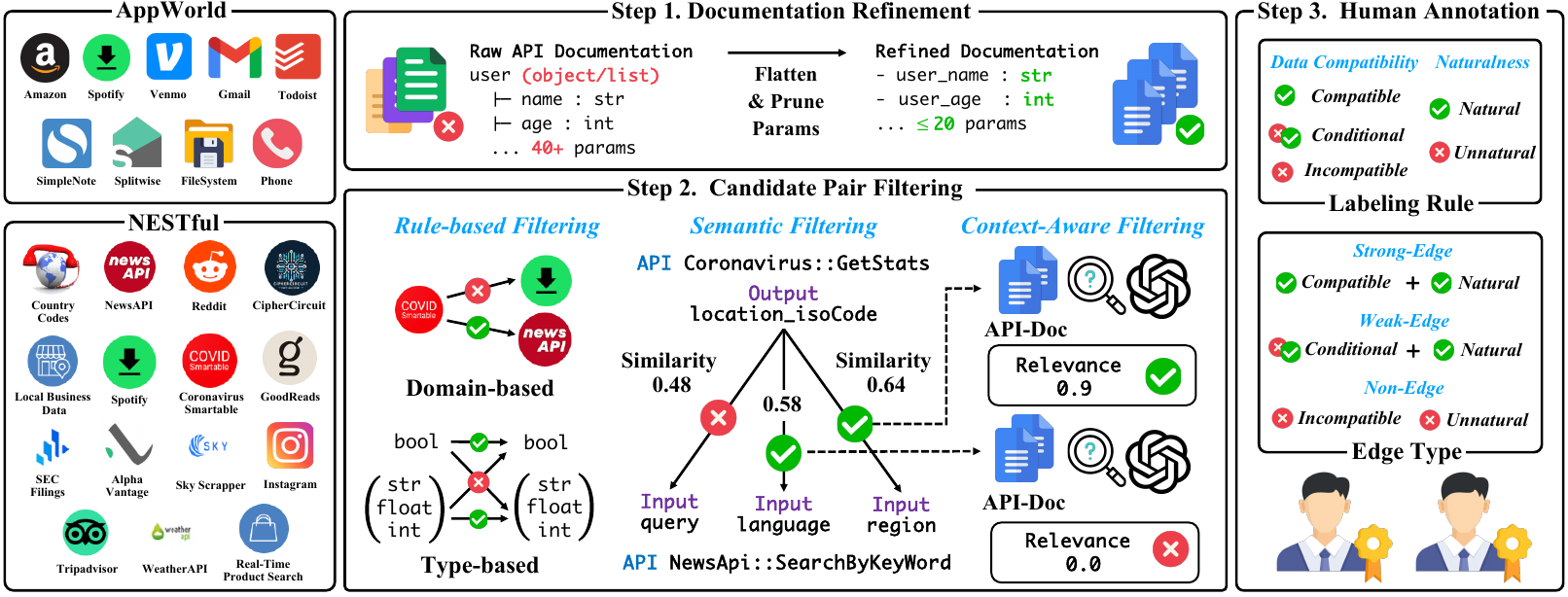}
  \caption{Construction process of the In-N-Out dataset: (1) refine API documentation, (2) filter candidate parameter pairs (rule-based, semantic, context-aware), (3) annotate edges by compatibility and naturalness.}
  \label{fig:dataset-construction}
\end{figure*}

To construct a parameter-level API graph, we aim to capture cases where an output parameter of one API is compatible with an input of another. Inferring such dependencies with LLMs is unreliable, since API documentations often contain ambiguities, vague references, and inconsistent terminology (see Appendix~\ref{appendix:ambiguous-api-documentation} for examples). Alternatively, one might attempt to verify dependencies through direct API execution, but this is often infeasible due to deprecated or inaccessible APIs, and execution success or failure does not guarantee true semantic compatibility.

Therefore, we adopt the perspective of a developer reading documentation to infer relationships between APIs, and use this as the basis for ground truth. Two experienced software developers, each with over five years of programming experience and a background in computer science, independently review API documentation, assess intended usage, and decide whether a valid connection exists between API parameters.

Our annotations are built on two realistic and diverse datasets: NESTful(v1) \cite{basu2025nestfulbenchmarkevaluatingllms} and AppWorld \cite{trivedi2024appworld}. NESTful contains 38 real-world APIs sourced from RapidAPI Hub\footnote{\url{https://rapidapi.com/hub}} across 15 domains (e.g., weather, music, news), along with 85 multi-tool queries. We extend it with 55 additional APIs from the same domains to enrich the graph with more diverse relationships. AppWorld offers a simulated environment for 9 real-world apps (e.g., amazon, venmo, spotify) with 457 APIs and 750 multi-tool tasks, providing a rich benchmark for compositional API usage. Based on these two datasets, we construct graphs that serve as a valuable resource for effective API composition.

We design a multi-stage pipeline to filter and annotate parameter-level connections between APIs. As shown in Figure~\ref{fig:dataset-construction}, the overall process consists of three steps: documentation refinement (Step 1), candidate pair filtering (Step 2), and human annotation (Step 3).

\subsection{Documentation Refinement}
\label{sec3:documentation-refinement}

API documentation varies widely in format, and some return outputs through object or list structures that bundle multiple pieces of information. Because our goal is to capture parameter-level relationships between APIs, we first flatten such structured outputs into individual parameters and retain only primitive types: bool, str, float, and int.

Real-world APIs often produce a large number of output parameters, many of which are auxiliary values that are unlikely to be connected to other APIs, such as execution status flags, image URLs, or timestamps. Including such rare connections in the graph would add noise when a tool agent later uses it. To avoid this, we prune each API’s outputs to at most 20 parameters, keeping only those that (1) support the API’s core functionality and (2) are plausible inputs to other APIs in realistic tool-use contexts.

Moreover, because many APIs provide only output schema examples without parameter descriptions, we provide the full API documentation to the GPT-4o mini model and instruct it to generate a concise description (see Appendix~\ref{appendix:prompt} for the prompt), which will later be used in the candidate filtering stage.

\subsection{Candidate Pair Filtering}
\label{sec3:candidate-pair-filtering}

Since the In-N-Out dataset defines edges at the parameter level, we first enumerate all possible pairs between output and input parameters. The number of initial pairs is extremely large: 279,928 for the 93 APIs in NESTful, and 1,916,351 for the 457 APIs in AppWorld. To reduce implausible connections and preserve the graph's utility, we apply three filtering processes---corresponding to Step~2 in Figure~\ref{fig:dataset-construction}.

\paragraph{Rule-based Filtering.} 
We eliminate pairs that are implausible either because their APIs belong to incompatible domains or because their types are mismatched. Domain-based filtering applies only to NESTful, which spans 15 diverse domains and thus produces irrelevant pairings such as combining a Coronavirus API with a Spotify API. Two experts independently reviewed the domain list and identified mutually incompatible domain pairs through agreement (see Appendix~\ref{appendix:domain-pairs-excluded} for the irrelevant domain pairs). For type compatibility, we group the four primitive types into two categories: (1) Boolean values (bool) and (2) textual/numeric values (str, int, float), allowing connections only within the same category.

\paragraph{Semantic Filtering.}
To capture more subtle mismatches, we use SBERT \cite{reimers2019sentence} to encode each parameter into an embedding based on its name and description, and filter pairs using cosine similarity. The similarity threshold is set to 0.5, chosen to ensure that all API parameter pairs appearing in the annotated call sequences of NESTful and AppWorld are retained. After this step, 73,526 candidate pairs remain in NESTful and 428,040 in AppWorld.

\paragraph{Context-aware Filtering.}
Embedding similarity alone may still preserve pairs that look related lexically but differ in practical meaning (e.g., location\_id vs. location\_name). To filter these cases, we prompt GPT-4o mini with the full documentation of both APIs and ask it to score the relevance of each pair on a 0–1 scale (see Appendix~\ref{appendix:prompt} for the prompt). Here, relevance refers to whether two parameters would reasonably be linked in real tool-use scenarios, beyond type or lexical similarity. Using the same ground-truth pairs described above to guide the cutoff, we set the threshold to 0.3, which retains all gold connections. After this step, 3,066 candidate pairs remain in NESTful and 48,757 in AppWorld.

To validate the filtering pipeline, we randomly inspected 100 discarded pairs from each dataset. In both NESTful and AppWorld, none of the sampled pairs represented connections that a developer would realistically use, suggesting that our filtering process effectively removes implausible edges while preserving the connections most useful for subsequent annotation and graph construction. (see Appendix~\ref{appendix:filtered-out-pairs} for the filtered-out pairs).

\subsection{Human Annotation}
\label{sec3:human-annotation}

To obtain reliable ground-truth labels, we conducted a human annotation process with the two experts introduced earlier on the parameter pairs that remained after filtering. Each candidate pair was evaluated according to two criteria: data compatibility and naturalness.

\textbf{Data compatibility} measures whether an output from one API can serve as a valid input for another. Pairs are labeled as \textit{compatible} if the output consistently provides the information required by the input (e.g., user\_id from one API used as user\_id in another). They are labeled as \textit{incompatible} if the output cannot fulfill the input requirement under any circumstance (e.g., an image URL cannot serve as a date input). The intermediate case, \textit{conditional}, applies if it depends on the content. For example, SimpleNote::ShowNote().content might contain an email address in some cases but not always, meaning the connection is possible but not guaranteed.

\textbf{Naturalness} evaluates whether the connection reflects realistic tool usage. For instance, ResetPassword() requiring a user's own email address is natural, but obtaining it through SearchFriends() would be contextually unnatural and deviated from typical API usage patterns, despite being technically possible. Pairs based on this criterion are labeled as \textit{natural} or \textit{unnatural}.

Before full-scale annotation, both annotators jointly labeled 100 parameter pairs to calibrate their understanding---ensuring a shared interpretations of the two criteria. Following this calibration, each annotator independently read the complete API documentation, including parameter names and descriptions, examined actual API call results if needed, and assigned labels. Pairs with any disagreement between the two annotators---9,271 pairs in AppWorld and 870 in NESTful---were resolved through discussion to produce final labels.

Based on the data compatibility and naturalness, each parameter pair was assigned to one of three edge types.

\textbf{Strong-Edge} occurs when the parameters are labeled as both compatible and natural, indicating a reliable and contextually appropriate data flow (e.g., Amazon::Login().access\_token → Amazon::ShowOrder().access\_token).

\textbf{Weak-Edge} corresponds to cases labeled as conditional and natural, where the connection is plausible but depends on certain conditions being met (e.g., SimpleNote::ShowNote().content → Gmail::SendEmail().email\_addresses).

\textbf{Non-Edge} is assigned to any pair involving incompatible or unnatural labels, reflecting parameter pairs that are not semantically aligned or executable in practice (e.g., Venmo::SearchUsers().last\_name → Phone::UpdateContact().email).

This process yields API graphs that reflect realistic and contextually valid parameter-level data flows for tool-agent reasoning.

\subsection{Data Statistics}
\label{sec3:data-statistics}

\begin{table}[t]
\centering
\resizebox{0.95\columnwidth}{!}{
\begin{tabular}{lcccc}
\toprule
\textbf{Dataset} & \textbf{Total Pairs} & \textbf{Annotated} & \textbf{Strong} & \textbf{Weak} \\
\midrule
NESTful   & 279,928   & 3,066   & 1,011 & 928 \\
\midrule
AppWorld  & 1,916,351 & 48,757  & 15,623 & 17,231 \\
\bottomrule
\end{tabular}
}
\caption{Statistics in the In-N-Out dataset: total parameter pairs, filtered human-annotated pairs, and resulting strong/weak edge counts.}
\label{tab:data-stats}
\end{table}

Table~\ref{tab:data-stats} presents the final annotation statistics. In the NESTful dataset, we identify 1,011 strong edges and 928 weak edges, together accounting for only 0.7\% of the 279,928 possible parameter pairs. In AppWorld, 15,623 strong edges and 17,231 weak edges constitute just 1.7\% of the 1,916,351 possible pairs. All remaining parameter pairs are marked as non-edges. This extreme sparsity highlights the difficulty of discovering compatible parameter connections without pre-constructed resources such as an API graph.

\begin{table}[t]
\centering
\resizebox{0.85\columnwidth}{!}{
\begin{tabular}{ccccc}
\toprule
\multirow{2}{*}{\textbf{Connection Level}} 
    & \multicolumn{2}{c}{\textbf{AppWorld}} 
    & \multicolumn{2}{c}{\textbf{NESTful}} \\
\cmidrule(lr){2-3} \cmidrule(lr){4-5}
 & \textbf{\(d_{avg}\)} & \textbf{Cross} & \textbf{\(d_{avg}\)} & \textbf{Cross} \\
\midrule
Param $\rightarrow$ Param  & 22 & 78\% & 7  & 63\% \\
API $\rightarrow$ API      & 32 & 76\% & 12 & 64\% \\
\bottomrule
\end{tabular}
}
\caption{
Connectivity statistics of In-N-Out. 
\(d_{avg}\) is the average number of incoming edges: (1) output parameters from other APIs linked to each input parameter (parameter level), and (2) upstream APIs providing inputs to each API (API level). 
Cross denotes the percentage of edges that cross domain boundaries.
}

\label{tab:connectivity-stats}
\end{table}

\begin{figure}[t]
  \centering
  \includegraphics[width=\columnwidth]{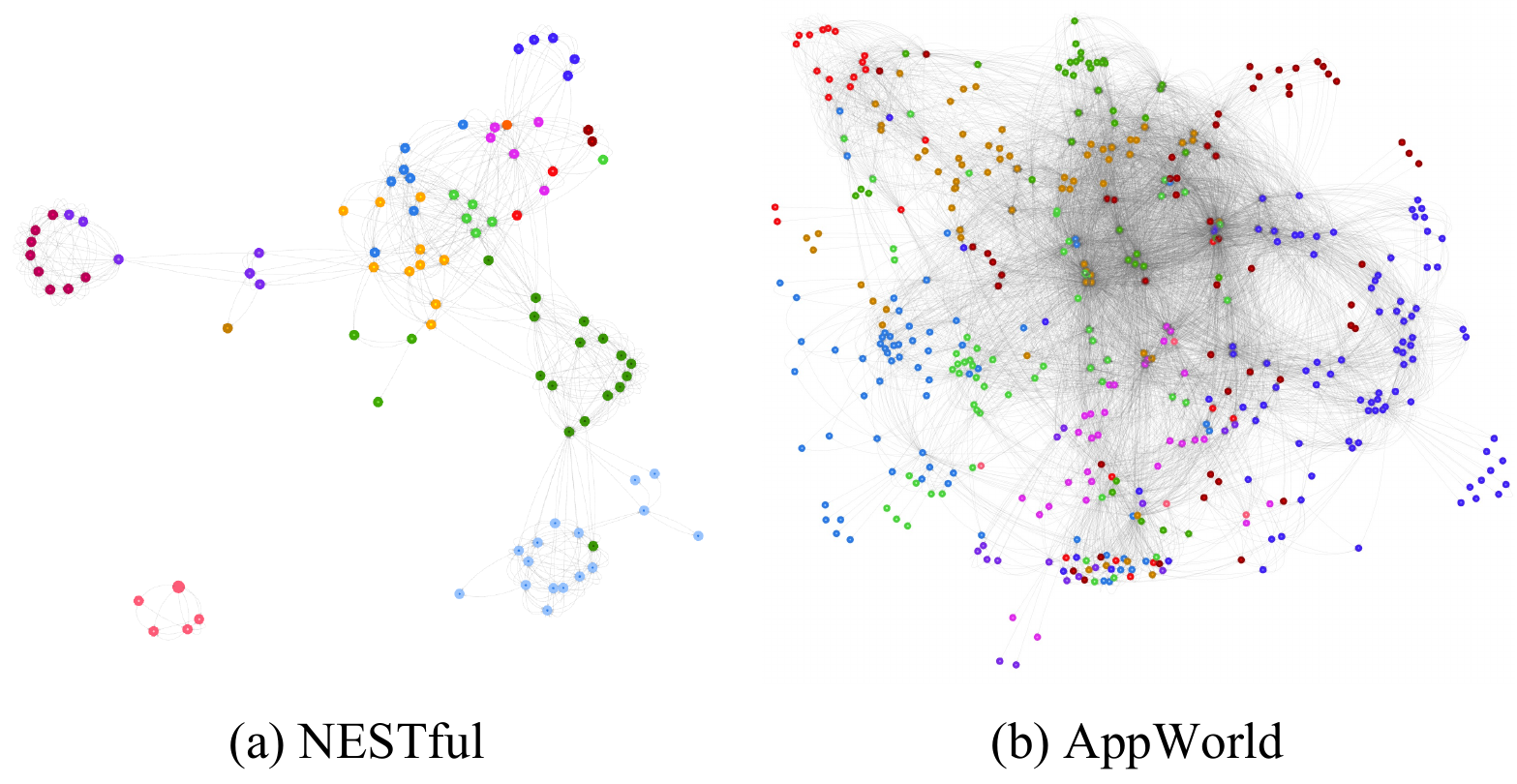}
  \caption{Gold API Graphs for (a) NESTful and (b) AppWorld datasets. Nodes with the same color belong to the same domain.}
  \label{fig:gold-api-graphs}
\end{figure}

As illustrated in Figure~\ref{fig:gold-api-graphs} and summarized in Table~\ref{tab:connectivity-stats}, the two datasets exhibit distinct connectivity patterns. In AppWorld, each API is connected to 32 other APIs on average, with 76\% of connections spanning across domains. In contrast, APIs in NESTful are connected to only 12 other APIs on average, with 64\% of them being cross-domain. At the parameter level, AppWorld also demonstrates relatively denser cross-domain links, whereas NESTful presents a more fragmented structure. By encompassing both styles, the In-N-Out dataset captures two representative patterns of API connectivity---one densely interconnected and the other more sparse---thus providing a robust resource for training and evaluating tool agents under diverse domain structures.

Although our graphs contain more cross-domain than in-domain connections (see the Cross column in Table~\ref{tab:connectivity-stats}), such multi-domain compositions are far less frequent in the original queries accompanied by NESTful and AppWorld. For example, only 22\% of NESTful queries and 30\% of AppWorld queries involve chaining APIs across different domains. This indicates that, without an explicit resource such as an API graph, it is particularly difficult to generate such cross-domain queries. Moreover, as we will show in the experiment section, agents can usually identify the prerequisite APIs needed for a target API call within the same domain, but their accuracy drops sharply across domains (around 80\% in-domain for both datasets vs.\ 38\% and 6\% cross-domain for NESTful and AppWorld, respectively). These findings highlight cross-domain reasoning as a major bottleneck for tool agents, and demonstrate that explicit parameter-level graphs are essential for guiding them toward the correct dependencies.

\section{Experiments}
\label{sec4:experiments}

Our experiments address two main questions: 
(1) can LLMs construct parameter-level API graphs comparable to human annotations, 
and (2) do such graphs improve downstream tool-agent performance? 
To answer these, we first benchmark LLMs on graph construction with and without fine-tuning on In-N-Out (\S{\ref{sec4:experiments-api-graph-construction}}). 
We then evaluate whether explicit API graphs help agents retrieve prerequisite APIs that supply inputs to another (\S{\ref{sec4:experiments-tool-retrieval}}), and select interdependent API subsets to support multi-tool query generation (\S{\ref{sec4:experiments-structured-api-subset-selection}}). We compare using gold graphs versus graphs generated by the fine-tuned LLM, quantifying the remaining gap due to imperfect graph construction. 
Finally, we conduct an end-to-end evaluation on $\tau$-bench (\S{\ref{sec4:experiments-end-to-end-evaluation}}) to examine whether the retrieval-level improvements observed in \S{\ref{sec4:experiments-tool-retrieval}} translate into better task completion in realistic, multi-turn tool-use settings.

\subsection{Benchmarking API Graph Construction Capabilities}
\label{sec4:experiments-api-graph-construction}

We first measure the zero-shot performance of various LLMs on the parameter-level API graph construction task, and then examine how much these models improve after fine-tuning on the In-N-Out. Since we build graphs for two distinct API datasets (NESTful and AppWorld), we also evaluate whether fine-tuning on one dataset generalizes effectively to the other.

\paragraph{Task Formulation.}
Given the full documentation $\mathcal{D}_A, \mathcal{D}_B$ of API $A$ and API $B$, along with a specific output parameter $p_{\text{out}}$ from $A$ and an input parameter $p_{\text{in}}$ from $B$, the model predicts the edge type---strong, weak, or non-edge:
\begin{equation}
f(\mathcal{D}_A, \mathcal{D}_B, p_{\text{out}}, p_{\text{in}}) \in \{\text{strong},\ \text{weak},\ \text{non}\}
\label{eq:edge-type-prediction}
\end{equation}
To increase task difficulty, we evaluate only parameter pairs that pass through all three filtering stages. From each dataset, we allocate 300 pairs for validation and 300 pairs for testing (100 per class in each split), and use the remaining pairs for training (2,466 pairs for NESTful and 48,157 for AppWorld). For a more rigorous evaluation, we ensure that APIs from two specific domains are excluded from training and included only in the test set.

\paragraph{Comparison Models.}
We evaluate three categories of models:  
(1) closed-source zero-shot LLMs---GPT-4o mini, GPT-4o, GPT-4.1 mini, GPT-4.1;  
(2) open-source zero-shot LLMs---Llama-3.2-3B, Llama-3.1-8B, Qwen2.5-7B, Qwen2.5-32B;  
(3) fine-tuned versions of the open-source models, trained on In-N-Out. We report both in-dataset results---training and evaluation on the same dataset---and cross-dataset results---training on one dataset and evaluating on the other. Detailed training settings are provided in Appendix~\ref{appendix:fine-tuning-setup}.

\paragraph{Results.}

\begin{table}[t]
\centering
\resizebox{0.9\columnwidth}{!}{
\begin{tabular}{l l c c}
\midrule
\multirow{2}{*}{\textbf{Model Type}} & \multirow{2}{*}{\textbf{Model}} & \multicolumn{2}{c}{\textbf{Accuracy (\%)}} \\
\cmidrule{3-4}
& & \textbf{NESTful} & \textbf{AppWorld} \\
\midrule
\multirow{4}{*}{Closed-source}
& GPT-4o mini    & 56.0 & 44.7 \\
& GPT-4o         & 61.0 & 42.3 \\
& GPT-4.1 mini   & 53.7 & \textbf{52.3} \\
& GPT-4.1        & \textbf{70.0} & 47.7 \\
\midrule
\multirow{4}{*}{Open-source}
& Llama-3.2-3B   & 39.0 & 41.3 \\
& Llama-3.1-8B   & 37.0 & 30.0 \\
& Qwen2.5-7B     & 47.7 & 39.7 \\
& Qwen2.5-32B    & \textbf{57.3} & \textbf{50.3} \\
\midrule
\multirow{4}{*}{\shortstack[l]{Fine-tuned\\(in-dataset)}}
& Llama-3.2-3B   & \textbf{76.7} & 93.0 \\
& Llama-3.1-8B   & 76.0 & 94.0 \\
& Qwen2.5-7B     & 74.0 & 91.0 \\
& Qwen2.5-32B    & 76.3 & \textbf{94.7} \\
\midrule
\multirow{4}{*}{\shortstack[l]{Fine-tuned\\(cross-dataset)}}
& Llama-3.2-3B   & 72.3 & 52.3 \\
& Llama-3.1-8B   & 70.7 & 61.3 \\
& Qwen2.5-7B     & 71.7 & 55.7 \\
& Qwen2.5-32B    & \textbf{74.3} & \textbf{66.3} \\
\midrule
\end{tabular}
}
\caption{Performance of different models in predicting parameter-level edge types in the API graph construction task across NESTful and AppWorld datasets.}
\label{tab:graph-construction-results}
\end{table}

Table~\ref{tab:graph-construction-results} summarizes the performance of all evaluated models. In the zero-shot setting, even high-performing closed-source models like GPT-4.1 reach only 70.0\% on NESTful and 47.7\% on AppWorld. Open-source models perform worse, with Qwen2.5-32B achieving 57.3\% on NESTful and 50.3\% on AppWorld. These results show that, despite their general reasoning ability, LLMs struggle to infer parameter-level API connections from documentation alone, highlighting the limitations of using zero-shot models for API graph construction.

Fine-tuning on In-N-Out yields large gains for all open-source models. Qwen2.5-32B reaches 76.3\% on NESTful and 94.7\% on AppWorld, while the smaller Llama-3.2-3B jumps from 39.0\% to 76.7\% and from 41.3\% to 93.0\%. These gains are achieved even though the test sets contain APIs from unseen domains. Notably, fine-tuning on AppWorld yields greater gains overall, likely because its graph contains nearly four times more APIs and over fifteen times more edges than NESTful---exposing models to a richer variety of parameter dependencies during training.

Cross-dataset experiments further confirm this generality. Qwen2.5-32B trained on NESTful and tested on AppWorld improves from 50.3\% to 66.3\%, surpassing the best zero-shot model. In the reverse setting, performance rises from 57.3\% to 74.3\%, nearly matching in-dataset fine-tuning. These results show that In-N-Out fine-tuning teaches general principles of parameter dependencies rather than dataset-specific patterns. We attribute this transferability to In-N-Out’s diverse coverage, enabling models to extend effectively to entirely new APIs beyond those in training.

\paragraph{Analysis.}
\label{sec4:analysis-on-api-graph-construction}
To further probe model limitations, we analyzed fine-tuned Qwen2.5-32B’s predictions in the cross-dataset setting, this time evaluating the model on the entire set of human-annotated parameter pairs. Specifically, we predicted edge types for all 48,757 annotated pairs in AppWorld and 3,066 in NESTful. The resulting automated graphs achieved 71.2\% accuracy on AppWorld and 71.5\% on NESTful. Since the labels are balanced across strong, weak, and non-edge classes, this evaluation reveals detailed error patterns.

\begin{figure}[t]
  \centering
  \includegraphics[width=\columnwidth]{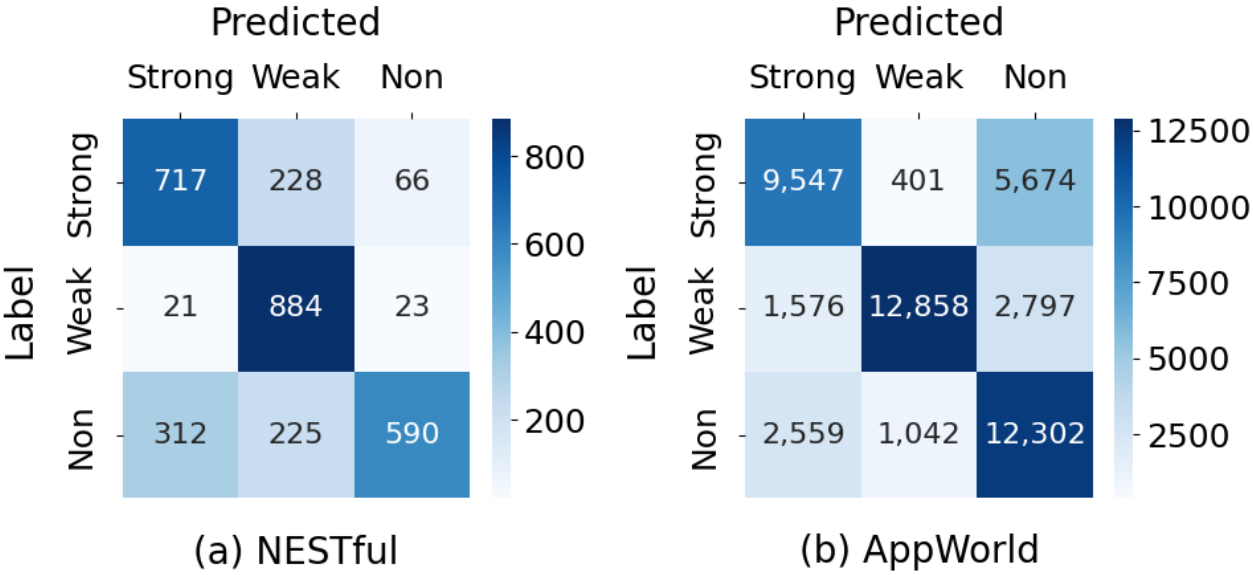}
  \caption{Confusion matrices of edge classification results with (a) NESTful, (b) AppWorld dataset.}
  \label{fig:confusion-matrices-of-edge-classification-results}
\end{figure}

Figure~\ref{fig:confusion-matrices-of-edge-classification-results} shows contrasting error trends across datasets. In AppWorld, the model tends to under-predict strong dependencies---misclassifying 46\% of cross-domain strong edges as non-edges, compared to 18\% for in-domain edges (e.g., Amazon::ShowOrder().delivery\_fee $\rightarrow$ Venmo::ShowTransactions().min\_amount).
In constrast, the model over-predicts in NESTful, labeling 48\% of in-domain and 24\% of cross-domain non-edges as strong (e.g., SkyScrapper::SearchAirport().entity\_id $\rightarrow$ SkyScrapper::GetPriceCalendar().sky\_id). These findings suggest that AppWorld errors mainly stem from difficulty in aligning semantically equivalent parameters across domains, whereas NESTful errors reflect over-reliance on lexical or structural similarity within a domain. Addressing these complementary failure modes will be key to improving the accuracy of automated graph construction.

\subsection{Tool Retrieval with API Graphs}
\label{sec4:experiments-tool-retrieval}

In realistic tool-use scenarios, solving a complex task often requires invoking multiple APIs in sequence, where each call may depend on outputs from earlier ones (Figure~\ref{fig:api-graph-usage}a). A key challenge is to identify which API can supply the missing input to a target API. This experiment evaluates whether API graphs from In-N-Out improve this retrieval process compared to a baseline without structural guidance.

\paragraph{Task Formulation.}
Given a task query $q$ and the documentation $\mathcal{D}_A$ of a target API $A$ with input parameters $\mathcal{P}(A)$, let $\mathcal{P}_{\text{miss}}(A) \subseteq \mathcal{P}(A)$ denote those parameters whose values are not explicitly specified in $q$. For each $p \in \mathcal{P}_{\text{miss}}(A)$, the objective is to select a prerequisite API $B$ whose output parameter can supply the value for $p$:
\begin{equation}
f(q, \mathcal{D}_A, p) \to B .
\label{equation:tool-retrieval}
\end{equation}

To simulate this process, we compute SBERT embeddings~\cite{reimers2019sentence} for all candidate APIs and rank them by cosine similarity to the description of $p$, as LLM-based ranking over the full API set is impractical due to context length limits. When an API graph is available, we re-rank candidates by first promoting graph-connected APIs to the top (preserving their similarity-based order), followed by the remaining unconnected ones. This strategy ensures that graph structure informs the ranking without discarding semantic relevance. The top-5 ranked candidates are then passed to GPT-4o mini, which selects the final API $B$ (see Appendix~\ref{appendix:prompt} for the prompt).

Evaluation instances are derived from task queries with gold API call sequences in the datasets. Each dependency where a target API requires an input provided by another API counts as a single instance, yielding 134 instances for NESTful. For AppWorld, 165 training and dev queries include gold sequences, but the 585 test queries do not. To obtain labels for these, we prompt GPT-4.1 to generate candidate sequences, execute them in the AppWorld simulator, and retain the 345 sequences that produce correct outputs (see Appendix~\ref{appendix:prompt} for the prompt). These were then validated by human annotators, resulting in 3,801 instances in total.

Performance is measured using:  
(1) Average Rank and Worst Rank of the correct API among all candidates ranked by embedding similarity;    
(2) Top-$k$ Accuracy for $k \in \{1, 2, 5, 10, 20\}$, i.e., the proportion of instances where the correct API appears among the top $k$;  
(3) Final Selection Accuracy, where GPT-4o mini chooses the correct API from the top-5 list.

\paragraph{Comparison Methods.}
We compare three retrieval settings. In the \textbf{no-graph} setting, all APIs are ranked purely by embedding similarity to the missing parameter description. The \textbf{gold graph} setting uses ground-truth edges from In-N-Out to guide re-ranking. The \textbf{automated graph} setting applies the same re-ranking strategy using edges predicted by a Qwen2.5-32B model that is fine-tuned on the other dataset (cross-dataset setup).

\paragraph{Results.}

\begin{table*}[t]
\centering
\resizebox{1.75\columnwidth}{!}{
\begin{tabular}{ll cc ccccc c}
\toprule
\multirow{2}{*}{\textbf{Dataset}} & \multirow{2}{*}{\textbf{Condition}} &
\multicolumn{2}{c}{\textbf{Rank (Gold API)}} &
\multicolumn{5}{c}{\textbf{Gold API included in Top-$k$ (\%)}} &
\multirow{2}{*}{\shortstack{\textbf{Final Selection}\\\textbf{Accuracy (\%)}}} \\
\cmidrule(lr){3-4}\cmidrule(lr){5-9}
 &  & \textbf{Avg.} & \textbf{Worst} & \textbf{Top-1} & \textbf{Top-2} & \textbf{Top-5} & \textbf{Top-10} & \textbf{Top-20} &  \\
\midrule
\multirow{3}{*}{\textbf{NESTful}}
& No-graph          & 3.1 & 13 & 43.3 & 56.0 & 83.6 & 97.8 & 100.0 & 68.7 \\
& Automated Graph   & 1.8 & 10 & 79.9 & 81.3 & 92.5 & 100.0 & 100.0 & 79.1 \\
& Gold Graph        & \textbf{1.6}  & \textbf{10}  & \textbf{84.3} & \textbf{89.6} & \textbf{92.5} & \textbf{100.0} & \textbf{100.0} & \textbf{79.9} \\
\midrule
\multirow{3}{*}{\textbf{AppWorld}}
& No-graph          & 19.3 & 449 & 51.8 & 55.0 & 58.6 & 63.9 & 68.7 & 57.3 \\
& Automated Graph   & 8.1 & 393 & 64.6 & 70.7 & 78.3 & 82.9 & 84.5 & 73.0 \\
& Gold Graph        & \textbf{4.5} & \textbf{393} & \textbf{75.4} & \textbf{81.1} & \textbf{88.7} & \textbf{94.0} & \textbf{95.7} & \textbf{82.8} \\
\bottomrule
\end{tabular}
}
\caption{Tool retrieval performance under three graph conditions on NESTful and AppWorld (using parameter descriptions as search queries).}
\label{tab:tool-retrieval_results}
\end{table*}

Table~\ref{tab:tool-retrieval_results} shows that incorporating API graphs significantly improves tool retrieval performance across both datasets. Notably, even graphs automatically constructed by fine-tuned Qwen2.5-32B yield substantial gains over the no-graph baseline, underscoring the value of structural information beyond embedding-based similarity.

On NESTful, rank-based metrics show a striking improvement when graphs are used. Compared to the no-graph baseline, the average rank of the correct API drops from 3.1 to 1.6 with the gold graph, and Top-1 accuracy rises from 43.3\% to 84.3\%, reducing the gap by over 40 percentage points. This confirms that retrieval is far more effective than embedding-based similarity alone once explicit graph structure is introduced. Moreover, even with automatically constructed graphs, performance closely approaches the gold graph; accordingly, Final Selection Accuracy also improves substantially, from 68.7\% (no-graph) to 79.1\% (automated) and 79.9\% (gold).  

In AppWorld, where the search space is much larger, the gains are even more pronounced. The average rank of the correct API improves from 19.3 without a graph to 4.5 with the gold graph, while Top-1 accuracy rises from 51.8\% to 75.4\%. Automated graphs again close much of this gap, reaching 8.1 average rank and 64.6\% Top-1 accuracy. Final Selection Accuracy also increases markedly, from 57.3\% (no-graph) to 73.0\% (automated) and 82.8\% (gold).

Overall, even imperfect automated graphs---constructed under the challenging cross-dataset setting---consistently outperform the no-graph baseline by re-ranking candidates so that the most relevant APIs appear near the top. These results highlight that more accurate graph construction would directly translate into further gains in retrieval accuracy, making graph-based retrieval even more powerful for tool agents.

\subsection{Structured API Subset Selection for Multi-Tool Query Generation}
\label{sec4:experiments-structured-api-subset-selection}

Generating multi-tool queries is essential not only for benchmarking but also for training models to solve complex tasks involving multiple APIs. A critical first step is to select subsets of APIs that can be composed into valid sequences via their input-output dependencies (Figure~\ref{fig:api-graph-usage}b). Here, we evaluate whether the In-N-Out graph helps models select such subsets more effectively. Specifically, we compare model performance with and without access to the graph, measuring how accurately LLMs identify API subsets that satisfy predefined structural patterns and target subset sizes.

\begin{figure}[t]
    \centering
    \includegraphics[width=\columnwidth]{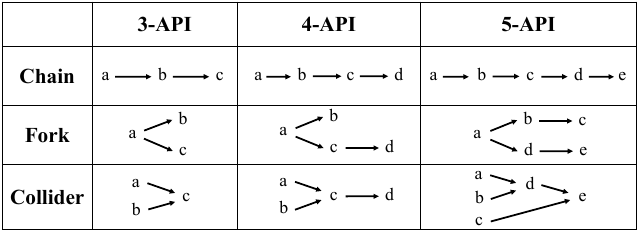}
    \caption{Predefined structural patterns (Chain, Fork, Collider) illustrated for 3-, 4-, and 5-API cases.}
    \label{fig:predefined-structural-patterns}
\end{figure}

\paragraph{Task Formulation.}
Given a set of APIs $\mathcal{A}$ with documentation $\mathcal{D}_{\mathcal{A}}$, the model must generate exactly five candidate subsets $\mathcal{S} = \{S_1, S_2, \dots, S_5\}$. Each $S_i$ contains $n \in \{3,4,5\}$ APIs whose parameter dependencies form a target structural pattern $\pi \in \Pi = \{\text{Chain}, \text{Fork}, \text{Collider}\}$ (Figure~\ref{fig:predefined-structural-patterns}). Formally, the task can be defined as:
\begin{equation}
f(\mathcal{D}_{\mathcal{A}}, n, \pi) \to \mathcal{S}, 
\quad |S_i| = n, \ |\mathcal{S}| = 5 .
\label{eq:pattern-subset-generation}
\end{equation}

For the 3-, 4-, and 5-API cases, $\mathcal{A}$ contains 15, 20, and 25 APIs respectively---sized to ensure that at least five valid subsets of the target pattern can be formed, while keeping the search space sufficiently challenging. Since additional valid connections may exist beyond the guaranteed ones, a generated subset is considered correct if it both satisfies the structural pattern and appears as a valid subgraph in the gold graph. Each configuration is sampled 100 times, and results are averaged across runs.

Performance is measured by precision: the proportion of generated subsets $S_i$ that meet the target pattern and appear in the gold graph, with duplicates counted once. As the model always outputs five subsets per instance, precision alone reliably reflects performance, without the need for recall.

\paragraph{Comparison Methods.}
We use GPT-4o mini as the base model and evaluate three conditions from \S{\ref{sec4:experiments-tool-retrieval}}. In the \textbf{no-graph} setting, the model selects subsets using only API documentation, with no structural guidance. In the \textbf{gold graph} setting, we additionally provide the ground-truth connections between the given APIs as part of the task instruction. In the \textbf{automated graph} setting, the same type of connection information is provided using edges predicted by the Qwen2.5-32B model fine-tuned on the other dataset (cross-dataset setup). Prompt details are provided in Appendix~\ref{appendix:prompt}.

\paragraph{Results.}

\begin{table}[t]
\centering
\footnotesize
\setlength{\tabcolsep}{4pt}
\renewcommand{\arraystretch}{0.9}
\resizebox{\columnwidth}{!}{%
\begin{tabular}{lllccc}
\toprule
\textbf{Dataset} & \textbf{\# APIs} & \textbf{Condition} & \textbf{Chain} & \textbf{Fork} & \textbf{Collider} \\
\midrule
\multirow{9}{*}{\textbf{NESTful}} 
& \multirow{3}{*}{3-API}
  & No-graph        & 58.2 & 38.6 & 38.8 \\
& & Automated Graph & 86.4 & 70.0 & 64.6 \\
& & Gold Graph      & \textbf{90.2} & \textbf{76.2} & \textbf{67.2} \\
\cmidrule(lr){2-6}
& \multirow{3}{*}{4-API}
  & No-graph        & 49.8 & 19.6 & 21.8 \\
& & Automated Graph & 76.6 & 23.6 & 41.2 \\
& & Gold Graph      & \textbf{84.0} & \textbf{34.4} & \textbf{43.8} \\
\cmidrule(lr){2-6}
& \multirow{3}{*}{5-API}
  & No-graph        & 37.2 & 17.4 & 6.4 \\
& & Automated Graph & 60.4 & 22.2 & 11.6 \\
& & Gold Graph      & \textbf{68.0} & \textbf{22.2} & \textbf{17.4} \\
\midrule
\multirow{9}{*}{\textbf{AppWorld}} 
& \multirow{3}{*}{3-API}
  & No-graph        & 41.2 & 32.2 & 23.8 \\
& & Automated Graph & 71.8 & 70.6 & 59.8 \\
& & Gold Graph      & \textbf{81.2} & \textbf{72.6} & \textbf{68.0} \\
\cmidrule(lr){2-6}
& \multirow{3}{*}{4-API}
  & No-graph        & 35.6 & 16.0 & 15.6 \\
& & Automated Graph & 47.0 & 34.0 & 41.0 \\
& & Gold Graph      & \textbf{60.8} & \textbf{40.2} & \textbf{46.4} \\
\cmidrule(lr){2-6}
& \multirow{3}{*}{5-API}
  & No-graph        & 5.6  & 13.0 & 6.2 \\
& & Automated Graph & 22.8 & 24.6 & 13.6 \\
& & Gold Graph      & \textbf{30.6} & \textbf{26.8} & \textbf{23.4} \\
\bottomrule
\end{tabular}%
}
\caption{Precision (\%) across API structure sizes and patterns under three graph conditions, for both NESTful and AppWorld.}
\label{tab:structured-api-subset-selection-results}
\end{table}

The results in Table~\ref{tab:structured-api-subset-selection-results} show that access to graph information significantly improves precision across structural patterns and subset sizes. For instance, in the 3-API chain setting, precision rises from 58.2\% to 90.2\% in NESTful and from 41.2\% to 81.2\% in AppWorld. Even in more challenging settings like the 5-API collider pattern, performance rises from 6.4\% to 17.4\% in NESTful and from 6.2\% to 23.4\% in AppWorld. These results demonstrate that gold graphs provide strong structural guidance that greatly enhances subset selection.

Automated graphs recover a large portion of this performance gap. In the AppWorld 3-API fork case, for example, precision reaches 70.6\% with the automated graph, recovering over 95\% of the gain achieved by the gold graph’s 72.6\%. Across conditions, automated graphs generally recover 70–90\% of the performance improvements provided by gold graphs, demonstrating that even imperfect structural cues can effectively guide LLMs in extracting valid API subsets.

When comparing datasets, we find that gold graphs yield consistently larger performance gains in AppWorld than in NESTful. This indicates that structural cues are especially helpful in more complex environments---like AppWorld, with denser connections and more cross-domain edges. As API sets grow larger and more entangled, graph-based supervision becomes increasingly essential for identifying valid API subsets beyond what documentation alone can support. Still, models struggle with more complex patterns (e.g., 5-API collider), underscoring the need to improve their ability to fully utilize structural information during compositional reasoning.

\subsection{End-to-End Performance Evaluation}
\label{sec4:experiments-end-to-end-evaluation}

While our earlier experiments (\S\ref{sec4:experiments-tool-retrieval}) demonstrate that API graphs substantially improve the retrieval of prerequisite APIs, this raises the question of whether such improvements translate into better end-to-end agent performance. To examine this, we conduct an additional evaluation using the $\tau$-bench~\cite{yao2024taubenchbenchmarktoolagentuserinteraction}, which evaluates multi-turn tool use in realistic user–agent dialogues.

\paragraph{Setup.}
$\tau$-bench consists of two domains, $\tau$-retail and $\tau$-airline, containing 15 and 13 API tools, respectively, and 115 and 50 test instances. Since the API documentation in $\tau$-bench does not include output parameter information, we manually created extended documentation by inspecting each API’s invocation results. From these, we enumerated all possible input–output parameter pairs---3,673 for $\tau$-retail and 7,033 for $\tau$-airline.

To construct API graphs, we used the Qwen2.5-32B model fine-tuned on the AppWorld portion of In-N-Out to predict edge types for all parameter pairs and retained only those classified as \textit{strong} edges. The resulting graphs contained 182 edges for $\tau$-retail and 537 edges for $\tau$-airline. Using these graphs, we augmented each API’s documentation by appending prerequisite API information to the corresponding input parameter descriptions. This allows the agent to reference dependency-aware documentation during tool planning and execution.

\paragraph{Comparison Methods.}
Following the experimental setup of $\tau$-bench, we adopted the tool calling agent described in the original paper as our base implementation. This agent uses function calling mode, where the model autonomously decides at each turn whether to respond to the user or invoke a tool. We compare two settings: the \textbf{baseline}, which uses the original $\tau$-bench documentation, and the \textbf{automated graph} setting, where prerequisite API information predicted from our constructed graphs are incorporated into each tool’s documentation. Both GPT-4o-mini and GPT-4o are used as the underlying reasoning models.

\paragraph{Results.}
Table~\ref{tab:tau-bench-results} summarizes the end-to-end task accuracy across both domains. Overall, incorporating graph information led to consistent improvements over the baseline, with the largest gain observed on $\tau$-airline using GPT-4o (from 50\% to 60\%). This indicates that structured dependency cues can help higher-capacity models utilize prerequisite information more effectively during tool planning.

Notably, when we qualitatively inspected the agent’s tool-calling trajectories, we observed that the graph-augmented documentation helped the agent correctly identify and invoke prerequisite APIs capable of supplying missing input values for subsequent calls. For instance, in $\tau$-airline baseline setting, the agent originally hallucinated a passenger’s \textit{dob} (date of birth) by substituting it with the flight’s departure date. After we added to the documentation that \textit{get\_user\_details} provides the required \textit{dob} parameter, the agent correctly called \textit{get\_user\_details} first and then used the retrieved date in the booking API, demonstrating the utility of structural guidance.

\begin{table}[t]
\centering
\resizebox{0.95\columnwidth}{!}{
\begin{tabular}{llcc}
\toprule
\multirow{2}{*}{\textbf{Model}} & \multirow{2}{*}{\textbf{Condition}} &
\multicolumn{2}{c}{\textbf{Accuracy (\%)}} \\
\cmidrule(lr){3-4}
 &  & \textbf{$\tau$-airline} & \textbf{$\tau$-retail} \\
\midrule
\multirow{2}{*}{GPT-4o-mini} 
& Baseline & 20.0 & 37.4 \\
& Automated Graph & \textbf{26.0} & \textbf{40.0} \\
\midrule
\multirow{2}{*}{GPT-4o}
& Baseline & 50.0 & 60.9 \\
& Automated Graph & \textbf{60.0} & \textbf{62.6} \\
\bottomrule
\end{tabular}
}
\caption{End-to-end task accuracy on $\tau$-bench. Results are reported for the tool calling agent using original documentation (Baseline) versus graph-augmented documentation (Automated Graph).}
\label{tab:tau-bench-results}
\end{table}

\paragraph{Discussion.}
Although graph augmentation led to measurable improvements, the overall gains were modest---a result that is closely reflects the inherent challenges of $\tau$-bench. According to the failure breakdown reported in the original paper, 56\% of errors stem from using wrong arguments or information, 25\% from incorrect decisions, and 19\% from partially resolved compound requests. Among these, only the first category can be partially addressed by the dependency structure encoded in our API graphs, whereas decision errors or domain-rule violations are largely orthogonal to graph-based reasoning. Moreover, a single task in $\tau$-bench can exhibit multiple failure types simultaneously, making overall end-to-end improvements difficult to achieve.  

Nevertheless, these findings suggest that while API graphs alone cannot resolve all reasoning failures, they meaningfully enhance the agent’s ability to provide correct inputs for dependent API calls. Future work may explore more sophisticated ways of integrating API graphs into the agent’s planning and dialogue process, beyond static documentation augmentation.
\section{Conclusion}
\label{sec5:conclusion}

In this paper, we introduce In-N-Out, a parameter-level API graph dataset that captures realistic input–output dependencies across APIs. By explicitly encoding these connections, our graphs help tool agents move beyond superficial heuristics and reason about how APIs can be composed. Experiments show that fine-tuning LLMs on In-N-Out significantly improves their ability to convert documentation into structured API graphs, narrowing the gap with expert-built graphs. As a result, agents equipped with In-N-Out achieve substantial gains in tool retrieval and structured API subset selection, demonstrating that explicit graph structure is a powerful lever for advancing tool-agent performance. We believe this structure could also enable future work on graph-based traversal strategies that plan optimal API sequences from given inputs to desired outputs.

\section{Limitations}
\label{sec6:limitations}

Our study has several limitations that suggest directions for future research. First, In-N-Out currently covers only two benchmark datasets. Expanding to broader domains would improve generalizability---particularly in heterogeneous or rapidly evolving tool ecosystems.

Second, some of our filtering and annotation criteria---such as judging whether a connection feels \textit{natural}---can be inherently subjective. While we aligned annotator criteria through joint calibration, future work could further decompose the notion of \textit{naturalness} into more objective dimensions. For instance, checking whether a user’s private information is being paired with another person’s public data can help flag unnatural information flow. Likewise, checking whether a parameter is being used across unrelated API domains can reveal domain-inappropriate usage. Such dimensions can provide clearer and more reproducible labeling guidelines.

Finally, our construction pipeline involves multiple filtering stages and substantial manual annotation. As the number of parameter pairs grows, the time and sustained attention required from annotators become a significant bottleneck. Future research could incorporate LLM-assisted labeling---for example, generating example input/output values to help annotators reason more intuitively about parameter compatibility. Complementary strategies such as retrieving similar previously labeled pairs may further support consistent decisions in ambiguous cases. In addition, a continual-learning workflow could reduce manual effort over time: a model trained on In-N-Out could propose labels for new parameter pairs, humans would verify them, and the refined annotations would be incorporated back into training. These directions would enable scalable dataset construction over a larger set of parameter pairs, which in turn may narrow the performance gap between automated and gold graphs and facilitate more robust tool-agent development.

\iftaclpubformat
\section*{Acknowledgments}
This work was supported by the National Research Foundation of Korea (NRF) grant (RS-2024-00333484) and by the Institute of Information \& Communications Technology Planning \& Evaluation (IITP) under the Leading Generative AI Human Resources Development grant (IITP-2026-RS-2024-00397085) and the grant (RS-2025-02215122, Development and Demonstration of Lightweight AI Model for Smart Homes), all of which are funded by the Korean government (MSIT). This work was also supported by the National Supercomputing Center with supercomputing resources including technical support (KSC 2024-CRE-0554, KSC-2025-CRE-0332).
\else
\fi

% \clearpage
% \bibliography{main}
\bibliographystyle{acl_natbib}

\clearpage
\appendix
\section*{Appendix}
\label{appendix}

\section{Examples of Ambiguous API Documentation}
\label{appendix:ambiguous-api-documentation}

\begin{lstlisting}[caption={Examples of ambiguous API documentation from NESTful dataset. Problematic parameters are highligheted in \textcolor{red}{red}.},  label={lst:challenging example nestful}, label={lst:ambiguous-documentation-example-nestful}, escapeinside=@@]
{
    "name": "Goodreads_GetBookByUrl",
    "description": "Get a book by its URL.",
    "query_parameters": [
        {
            "name": "bookURL",
            "type": "str",
            "description": "The URL of the book on Goodreads."
        }
    ],
    "output_parameters": [
        "@\textcolor{red}{author\_id}@": "str",
        "@\textcolor{red}{author\_legacyId}@": "int",
       ...
    ]
},
{
    "name": "Goodreads_GetAuthorsBooks",
    "description": "Retrieves books authored by a specific author.",
    "query_parameters": [
        {
            "name": "@\textcolor{red}{authorID}@",
            "type": "str",
            "description": "ID of the author whose books are to be retrieved. Can be obtained through other endpoints."
        },
        ...
    ],
    "output_parameters": [
        "bookId": "str",
        ...
    ]
}
\end{lstlisting}

In Listing~\ref{lst:ambiguous-documentation-example-nestful}, 
Goodreads\_GetBookByUrl returns two identifiers (author\_id, author\_legacyId), while Goodreads\_GetAuthorsBooks requires a single authorID. The documentation does not clarify whether authorID corresponds to one or both of the returned identifiers, 
making the intended mapping ambiguous.

\vspace{0.5em}
\begin{lstlisting}[caption={An example of ambiguous API documentation from AppWorld dataset. Problematic parameters are highlighted in \textcolor{red}{red}.}, label={lst: LMSYS license terms}, label={lst:ambiguous-documentation-example-appworld}, escapeinside={(*@}{@*)}]
{
    "app_name": "Phone",
    "api_name": "SearchContacts",
    "description": "Search your contact book for relatives' information.",
    "parameters": [
        {
            "name": "(*@\textcolor{red}{query}@*)",
            "type": "string",
            "description": "Search query for the contacts list."
        },
        {
            "name": "(*@\textcolor{red}{relationship}@*)",
            "type": "string",
            "description": "Relationship with the person in the contacts list to filter by."
        },
        ...
    ],
    "response_schemas": {
        "success": [
            {
                "contact_id": 1,
                "first_name": "string",
                "last_name": "string",
                "email": "user@example.com",
                ...
            }
        ],
        "failure": {
            "message": "string"
        }
    }
}
\end{lstlisting}

Listing~\ref{lst:ambiguous-documentation-example-appworld} shows a case from AppWorld. The SearchContacts API accepts both query and relationship, but the documentation is unclear: it is unspecified whether query accepts only names or also phone numbers/emails, and whether relationship terms should be passed in query or separately in relationship. Such ambiguities complicate reliable inference of parameter-level connections.

\section{Domain Pairs Excluded by Rule-based Filtering}
\label{appendix:domain-pairs-excluded}

\begin{figure}[t]
  \centering
  \includegraphics[width=\columnwidth]{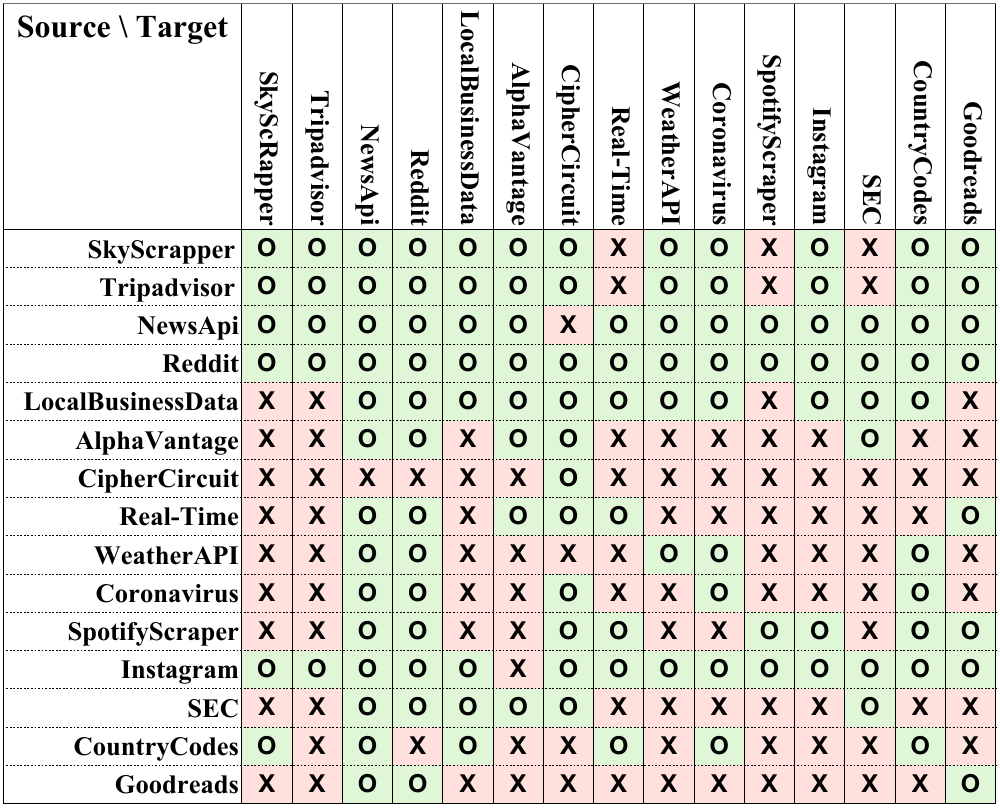}
  \caption{Irrelevant Domain Pairs for Domain Filtering list. Domain pairs are filtered by \emph{Source$\rightarrow$Target} flow, retaining only cases where a source API’s output can serve as a target API’s input.}
  \label{fig:irrelevant-domain-pairs}
\end{figure}

Figure~\ref{fig:irrelevant-domain-pairs} shows the domain pairs excluded during rule-based filtering. For NESTful, which spans 15 heterogeneous domains, we remove implausible combinations such as \emph{Coronavirus$\rightarrow$Spotify}. Two experts independently reviewed the domain list and agreed on mutually incompatible domain pairs, ensuring that only source–target pairs with potential parameter-level connections remain in the dataset.

\section{Fine-tuning Setup for API Graph Construction on In-N-Out}
\label{appendix:fine-tuning-setup}

We fine-tuned models on the In-N-Out dataset to improve API graph construction accuracy. 
For NESTful, training was run for 30 epochs, while AppWorld was trained for 10 epochs. 
Checkpoints were selected based on validation performance to avoid overfitting. 
Listing~\ref{lst:fine-tuning-config} summarizes the core hyperparameters used in our LoRA-based fine-tuning setup. 
For Llama-3.2-3B, Llama-3.1-8B, Qwen2.5-7B, and Qwen2.5-32B, the best validation epochs were 20/18/26/28 for NESTful, 
and 3/1/2/3 for AppWorld, respectively.

\begin{lstlisting}[caption={Fine-tuning configuration details.},  label={lst:fine-tuning-config}, escapeinside=@@]
gradient_accumulation_steps: 16
max_grad_norm: 1.0
optimizer:
  type: Adam8bit
  params:
    learning_rate: 1.0e-5
    betas: [0.9, 0.995]
    weight_decay: False
LoRA Finetuning (PEFT):
  lora_alpha: 16
  lora_dropout: 0.1
  lora_r: 64
\end{lstlisting}

\section{Examples of Filtered-out Pairs}
\label{appendix:filtered-out-pairs}

Shown below are pairs removed by Semantic Filtering: two NESTful examples followed by two AppWorld examples.
\vspace{0.5em}

\apimapboxsim{Instagram::Following().\zw profile\_pic\_url}
             {LocalBusinessData::BusinessReviews().\zw language}
             {0.32}
             
\vspace{0.5em}

\apimapboxsim{SEC::CashFlows().\zw filing\_filingDate}
             {Tripadvisor::SearchHotels().\zw sort}
             {0.48}
\vspace{0.5em}

\apimapboxsim{Phone::ShowAlarms().\zw repeat\_days}
             {Venmo::ShowAccount().\zw access\_token}
             {0.35}
             
\vspace{0.5em}

\apimapboxsim{Todoist::ShowTask().\zw section\_id}
             {Gmail::MarkThreadStarred().\zw email\_thread\_id}
             {0.43}
\vspace{0.5em}
             
Shown below are pairs removed by Context Aware Filtering: two NESTful examples followed by two AppWorld examples.
\vspace{0.5em}

\apimapbox{Instagram::\zw SearchReels().\zw user\_id}
          {SpotifyScraper::\zw GetTrackMetadata().\zw trackId}
          {0.0}

\vspace{0.5em}

\apimapbox{Reddit::PostsByUsername().\zw post\_subreddit}
          {SpotifyScraper::GetTrackMetadata().\zw trackId}
          {0.0}

\vspace{0.5em}

\apimapbox{Amazon:\zw SearchProducts().\zw variations\_color}
          {Splitwise::\zw UpdatePaymentComment().\zw comment}
          {0.2}

\vspace{0.5em}

\apimapbox{FileSystem::ShowAccount().\zw last\_logged\_in}
          {venmo::DownloadBankTransferReceipt().\zw access\_token}
          {0.0}

\section{Prompt Templates}
\label{appendix:prompt}

The following are all prompt templates used in our study:
\begin{itemize}
  \item Documentation Refinement (\S{\ref{sec3:documentation-refinement}}): Figure~\ref{tab:prompt-template-documentation-refinement}.
  \item Context-Aware Filtering (\S{\ref{sec3:candidate-pair-filtering}}): Figure~\ref{tab:prompt-template-context-aware-filtering}.
  \item API graph Construction (\S{\ref{sec4:experiments-api-graph-construction}}): Figure~\ref{tab:prompt-template-api-graph-construction}.
  \item Generating API call sequences for AppWorld Test Set (\S{\ref{sec4:experiments-tool-retrieval}}): We adopt the original ReAct prompt from~\cite{trivedi2024appworld}.
  \item Tool Retrieval with API Graphs (\S{\ref{sec4:experiments-tool-retrieval}}): Figure~\ref{tab:prompt-template-tool-retrieval}.
  \item Structured API Subset Selection for Multi-Tool Query Generation (\S{\ref{sec4:experiments-structured-api-subset-selection}}): Figure~\ref{tab:prompt-template-structed-api-subset-selection}.
\end{itemize}

\begin{table*}[t!]
    \centering
    \small
    \renewcommand{\arraystretch}{1.5}
    \begin{tabular}{p{0.95\textwidth}}
        \Xhline{2\arrayrulewidth}
        \textbf{Documentation Refinement} \\
        \hline
        You are a helpful assistant for a developer who is trying to understand the usage of an API.\\
        The developer will provide you with the API name, description, input parameters, and output parameters. Your job is to generate a description for specific output parameters of the API. \\
        The description must be concise, and help the developer understand the information that the output parameter provides. Do not generate information that is not present in the API documentation.\\
        API: \textcolor{blue}{\{api\_documentation\}}\\
        I need a description for the output parameter \textcolor{blue}{\{output\_parameter\_name\}}.\\
        \Xhline{2\arrayrulewidth}
    \end{tabular}
    \caption{Prompt template for Documentation Refinement task (\S{\ref{sec3:documentation-refinement}}).}
    \label{tab:prompt-template-documentation-refinement}
\end{table*}

\begin{table*}[t!]
    \centering
    \small
    \renewcommand{\arraystretch}{1.5} % Adjust row height
    \begin{tabular}{p{0.95\textwidth}}
        \Xhline{2\arrayrulewidth}
        \textbf{API Graph Constrcution} \\ 
        \hline
        You are a helpful assistant that classifies whether a connection (edge) between two API parameters is valid and natural.\newline

        You will be given:\newline
        -- A source API (with an output parameter)\newline
        -- A target API (with an input parameter)\newline

        [Criteria]\newline
        1.\ Data Compatibility: Can the source output consistently be used as the target input (i.e., do they refer to the same kind of entity/information)?\newline
        \quad-- Always: classify as ``compatible''.\newline
        \quad-- Only in some cases: classify as ``conditional''.\newline
        \quad-- Not at all: classify as ``incompatible''.\newline
        2.\ Naturalness: Is it natural, based on common user behavior, to pass the source output into the target input?\newline
        \quad-- Natural: ``natural''. \quad\newline
        -- Not natural: ``unnatural''.\newline

        [Edge label]\newline
        -- compatible \& natural $\Rightarrow$ ``strong-edge''\newline
        -- conditional \& natural $\Rightarrow$ ``weak-edge''\newline
        -- otherwise $\Rightarrow$ ``non-edge''\newline

        [Output format (single JSON object)]\newline
        \{\newline 
        \hspace*{2em}``source\_api'': ``name\_of\_source\_api'',\newline 
        \hspace*{2em}``target\_api'': ``name\_of\_target\_api'',\newline 
        \hspace*{2em}``source\_param'': ``name\_of\_source\_output\_param'', \newline
        \hspace*{2em}``target\_param'': ``name\_of\_target\_input\_param'', \newline
        \hspace*{2em}``edge\_type'': ``strong-edge'' $\vert$ ``weak-edge'' $\vert$ ``non-edge''\newline 
        \}\newline

        Now, classify the following edge:\newline
        Source API Information: \textcolor{blue}{\{source\_api\_documentation\}}\newline
        Target API Information: \textcolor{blue}{\{target\_api\_documentation\}}\newline
        Edge to classify: \textcolor{blue}{\{source\_parameter\_name\}} $\rightarrow$ \textcolor{blue}{\{target\_parameter\_name\}}
        \\
        \Xhline{2\arrayrulewidth}
    \end{tabular}
    \caption{Prompt template for API Graph Construction task (\S{\ref{sec4:experiments-api-graph-construction}})}
    \label{tab:prompt-template-api-graph-construction}
\end{table*}

\begin{table*}[t!]
    \centering
    \small
    \renewcommand{\arraystretch}{1.5} % Adjust row height
    \begin{tabular}{p{0.95\textwidth}}
        \Xhline{2\arrayrulewidth}
        \textbf{Context-Aware Filtering} \\ 
        \hline
        %\textbf{<Prompt>}\newline
        You are an assistant who helps developers understand the relevance between two APIs.\\
The developer will provide you with the documentation of the APIs.  
One is the source API, and the other is the target API. Your job is to determine if the specified output parameter of the source API can give full information to the specified input parameter of the target API. \\
Then you need to generate a relevance score between 0 and 1, which indicates the probability of the output parameter being relevant to the input parameter.\newline

There are some rules you should consider:

1. Prioritize parameter descriptions: Do not rely solely on parameter names to infer their meaning. Always refer to the parameter description, as names can be misleading.

2. Complete information matching: A source parameter can be linked to multiple target parameters if it contains all necessary information. However, partial matches are not allowed.

3. Type compatibility is flexible: Parameter types do not have to be identical as long as the value meaning is preserved.\newline

Source API : \textcolor{blue}{\{source\_api\_documentation\}}\newline
Target API : \textcolor{blue}{\{target\_api\_documentation\}}\newline
Source Parameter: \textcolor{blue}{\{source\_parameter\_information\}}\newline
Target Parameter:\textcolor{blue}{\{target\_parameter\_information\}}
        \\
        \Xhline{2\arrayrulewidth}
    \end{tabular}
    \caption{Prompt template for Context-Aware Filtering task (\S{\ref{sec3:candidate-pair-filtering}})}
    \label{tab:prompt-template-context-aware-filtering}
\end{table*}

\begin{table*}[t!]
    \centering
    \small
    \renewcommand{\arraystretch}{1.5} % Adjust row height
    \begin{tabular}{p{0.95\textwidth}}
        \Xhline{2\arrayrulewidth}
        \textbf{Tool Retrieval} \\ 
        \hline
        %\textbf{<Prompt>}\newline
       You are a helpful assistant that retrieves relevant APIs to obtain a specific input parameter for a given API.

- You are given a task instruction, the documentation of a primary API, and a list of additional available APIs.\newline
- You are also given the name and description of a specific input parameter for the primary API.\newline
- Your goal is to determine how to obtain the value for this input parameter, using the available APIs if necessary.\newline

Your output should be in the following JSON format:\newline
\{\newline
\hspace*{2em}``observation'': ``A brief description of how you can obtain the parameter.'',\newline
\hspace*{2em}``prerequisite\_api'': ``The name of the API you need to call to obtain this parameter, if applicable.'' \newline
\}\newline

Instructions:\newline
- If the parameter can be directly inferred from the task instruction or assumed as a default, explain this in `observation' and leave `prerequisite\_api' as an empty string.\newline
- If the parameter requires calling another API to obtain, describe this in `observation' and specify the required API in `prerequisite\_api'.\newline

Now, please analyze the task instruction, the primary API's documentation, the target input parameter, and the list of additional available APIs.\newline

Task Instruction: \textcolor{blue}{\{task\_instruction\}}\newline
Primary API Documentation: \textcolor{blue}{\{api\_documentation\}}\newline
Target Input Parameter: \textcolor{blue}{\{target\_parameter\_name\}}\newline
Additional APIs available: \textcolor{blue}{\{top-5 relevant\_api\_documentation\}}
        \\
        \Xhline{2\arrayrulewidth}
    \end{tabular}
    \caption{Prompt template for Tool Retrieval task (\S{\ref{sec4:experiments-tool-retrieval}})}
    \label{tab:prompt-template-tool-retrieval}
\end{table*}

\begin{table*}[t!]
    \centering
    \small
    \renewcommand{\arraystretch}{1.5} % Adjust row height
    \begin{tabular}{p{0.95\textwidth}}
        \Xhline{2\arrayrulewidth}
        \textbf{Structured API Subset Selection for Multi-Tool Query Generation} \\ 
        \hline
        %\textbf{<Prompt>}\newline
        You are a helpful assistant that selects small sets of APIs that match a strict dependency pattern.\newline

[Pattern]\newline
- APIs: 1,2,3,4,5\newline
- Edges (must be satisfied internally): \textcolor{blue}{\{EDGE\_LINES: 1$\to$2, 2$\to$3, 3$\to$4, 4$\to$5\}}\newline

[Task]\newline
From api\_list\_block, identify up to 5 distinct valid groups of 5 APIs that satisfy the above pattern.
In this pattern, an edge is valid only if an output parameter of one API provides the full information required by the corresponding input parameter of another API. Do not output the pattern or any explanation — only the group mappings.\newline

{\color{gray!100}%
[Connections]\newline
You are provided with connection\_list, which are known helpful relationships.\newline
However, it is not mandatory that all required edges appear there. You may also check input-output compatibility.%
}\newline

[Output Format]\newline
Output only in the following format (for each valid group):\newline
APIs: 1: <API 1>, 2: <API 2>, 3: <API 3>, 4: <API 4>, 5: <API 5>\newline

[Rules]\newline
- Replace <API num> with the exact API names from api\_list\_block.\newline
- Keep the numbering 1..5 exactly as shown.\newline
- Separate multiple groups with a line containing only `---'.\newline
- Do not output placeholders like ``app1.api1''; use real API names only.\newline
- Do not add extra commentary, JSON, or text outside the specified block.\newline
- Do not repeat identical (1..5) groups.\newline

api\_list\_block: \textcolor{blue}{\{api\_list\_block\}}\newline
{\color{gray!100} connection\_list: \textcolor{blue}{\{connection\_list\}}}
        \\
        \Xhline{2\arrayrulewidth}
    \end{tabular}
    \caption{Prompt template for Structured API Subset Selection for Multi-Tool Query Generation. In the no-graph setting, experiments were conducted with the gray text omitted. (\S{\ref{sec4:experiments-structured-api-subset-selection}})}
    \label{tab:prompt-template-structed-api-subset-selection}
\end{table*}

\end{document}